\pdfoutput=1
\documentclass[a4paper]{article}
\usepackage{arxiv}
\usepackage[T1]{fontenc}
\usepackage[utf8]{inputenc} %support umlauts in the input
\usepackage[%
  rm={oldstyle=false,proportional=true},%
  sf={oldstyle=false,proportional=true},%
  tt={oldstyle=false,proportional=true,variable=false},%
  qt=false%
]{cfr-lm}
\usepackage{graphicx}
\usepackage[%
  square,        % for square brackets
  comma,         % use commas as separators
  numbers,       % for numerical citations;
  sort&compress, % as sort but in addition multiple numerical citations
]{natbib}
\RequirePackage[%
  final,%
  expansion=alltext,%
  protrusion=alltext-nott]{microtype}%
\SetExpansion
[ context = sloppy,
  stretch = 30,
  shrink = 60,
  step = 5 ]
{ encoding = {OT1,T1,TS1} }
{ }
\usepackage{amssymb}
\usepackage{mathtools}
\usepackage{stmaryrd}
\renewcommand{\vec}[1]{\boldsymbol{#1}}  % Bold characters for vectors

\usepackage[caption=false]{subfig}

\usepackage[group-four-digits,per-mode=fraction]{siunitx}

\usepackage{booktabs}

\usepackage{paralist}

\usepackage{hyperref}
\hypersetup{hidelinks,
  colorlinks=true,
  allcolors=black,
  pdfstartview=Fit,
  breaklinks=true}
\usepackage[all]{hypcap}
\usepackage[capitalise,nameinlink]{cleveref}
\crefname{section}{Section}{Section}
\Crefname{section}{Section}{Sections}

\newcommand{\IR}{\mathbb{R}}
\newcommand{\IN}{\mathbb{N}}

\newcommand{\cX}{\mathcal{X}}
\newcommand{\cZ}{\mathcal{Z}}
\newcommand*\from{\colon}

\newcommand{\ie}{i.\,e.,\ }

\DeclarePairedDelimiter{\norm}{\lVert}{\rVert}
\DeclarePairedDelimiter{\abs}{\lvert}{\rvert}

\usepackage{authblk}

\author[1\thanks{\tt{kiudee@mail.upb.de}}]{Karlson Pfannschmidt}
\author[1]{Eyke Hüllermeier}
\affil[1]{Heinz Nixdorf Institute, Paderborn University, Paderborn, Germany}

\begin{document}
\title{Learning Choice Functions via Pareto-Embeddings%
\thanks{Preprint of an article presented at KI 2020, 43. German Conference on Artificial Intelligence, Bamberg, Germany}}
%
%\titlerunning{Abbreviated paper title}
% If the paper title is too long for the running head, you can set
% an abbreviated paper title here
%
%\author{Karlson Pfannschmidt\inst{1}\orcidID{0000-0001-9407-7903} \and Eyke Hüllermeier\inst{1}\orcidID{0000-0002-9944-4108}}
%\author{\href{https://orcid.org/0000-0001-9407-7903}{\includegraphics[scale=0.06]{pics/orcid}\hspace{1mm}Karlson~Pfannschmidt}
%\And
%\href{https://orcid.org/0000-0002-9944-4108}{\includegraphics[scale=0.06]{pics/orcid}\hspace{1mm}Eyke~Hüllermeier}\\
%\And \\
%Department of Computer Science
%}

%
%\authorrunning{K. Pfannschmidt and E. Hüllermeier}
% First names are abbreviated in the running head.
% If there are more than two authors, 'et al.' is used.
%
%\institute{Paderborn University, Germany}
%
\maketitle              % typeset the header of the contribution
\begin{abstract}
  We consider the problem of learning to choose from a given set of objects, where each object is represented by a feature vector.
  Traditional approaches in choice modelling are mainly based on learning a latent, real-valued utility function, thereby inducing a linear order on choice alternatives.
  While this approach is suitable for discrete (top-1) choices, it is not straightforward how to use it for subset choices.
  Instead of mapping choice alternatives to the real number line, we propose to embed them into a higher-dimensional utility space, in which we identify choice sets with Pareto-optimal points.
  To this end, we propose a learning algorithm that minimizes a differentiable loss function suitable for this task.
  We demonstrate the feasibility of learning a Pareto-embedding on a suite of benchmark datasets.

\end{abstract}
\section{Introduction}
% What are choice functions and why is it an interesting problem to learn these?
The quest for understanding and modeling human decision making  has a long history
in various scientific disciplines, including economics and psychology
\cite{Domshlak}.
Starting with the seminal work by \citet{arrow1951},
\emph{choice functions} have been analyzed as a key concept of a formal theory
of choice.
In simple terms, a decision maker is confronted with a (possibly varying) set
of alternatives and the choices made are observed.
The ultimate goal is to explain and predict the choice behavior.

% What is the learning problem we try to solve?
In machine learning, the task of ``learning to choose'' is part of the broader field of
\emph{preference learning}, which attracted increased attention in recent years
\cite{PL-book}.
% also due to its applicability in information retrieval and eCommerce.
The task for a learner is to observe choices from multiple sets of objects,
and to produce a function which maps from candidate sets to choice sets.
%Typically, each object is represented by a real-valued vector of features.
An important special case is the setting in which the decision maker only
chooses one object from each given set, which is known as \emph{discrete choice}.
A popular strategy to tackle the learning problem is to posit that the
choice probabilities depend on an underlying real-valued utility function of the
decision maker.
Under this assumption, learning can be accomplished by identiying the parameters of such a function.
% TODO: What are the origins of this problem and how was it solved so far?
The more general problem of predicting choices in the form of \emph{subsets} of objects has
been considered only very recently \cite{benson2018,pfannschmidt2019}.
Extending the approach based on utility functions  toward this setting turns out to be non-trivial.
Either one faces combinatorial problems calculating the probabilities for many
subsets \cite{benson2018}, or has to resort to thresholding techniques \cite{pfannschmidt2019}.

We propose to solve this problem by embedding the objects in a
higher-dimensional utility space, in which subset choices are naturally identified by Pareto-optimal points (\cref{sec:pareto_embedding}).
To learn a suitable embedding function, we devise a differentiable loss
function tailored to this task.
% (\cref{sub:learning}).
We then utilize the loss function as part of a deep learning pipeline to
investigate the feasibility of learning such a Pareto-embedding
(\cref{sec:evaluation}).

\section{Modeling Choice}\label{sec:choice}

We proceed from a reference set of objects (choice alternatives) $\cX \subset \IR^d$, which,
for ease of exposition, is assumed to be finite.
Each $\vec{x} \in \cX$ is represented as a vector of real-valued features
$(x_1, \dots, x_d)$.
We call a finite subset of objects $Q \subseteq \cX$
a \emph{choice task} and allow the size $|Q| \in \IN$ to vary across tasks.
For each choice task $Q = \{ \vec{x}_1, \ldots, \vec{x}_m \}$, we assume that a preference is expressed in terms
of a \emph{choice set} $C \subseteq Q$.
% TODO (if enough space): bipartition
%We call $\cC = \bigcup_{n\in\IN} \cC_n$ the \emph{choice space}, where $\cC_n$ is the set of all subsets of $[n] \coloneqq \{1, 2, \dots, n\}$.
A useful representation of a choice set is in terms of a binary vector
$\vec{c} \in \{0, 1\}^m$, where $c_i = 1$ if $\vec{x}_i \in C$ and $c_i = 0$ if $\vec{x}_i\not\in C$.
%---and similarly, $\cC_m = \{0, 1\}^m$.
%We will use both representations interchangeably.

% TODO: Introduce a subsection only if needed:
%\subsection{Utility-Based Choices}
%\label{sub:utility_choices}

One of the first approaches to explaining choices was to assume that a decision maker
can assign a (latent) utility to each of the choice alternatives.
Formally, we represent such a utility function as a function
$
  \cX \to \IR
$
from the space of objects to the real numbers.
Based on these utilities, a rational decision maker will always pick the
alternative with the highest utility, \ie the top-1 object.
To explain variability in choices, noise can be  added to the utilities, which
results in what is called a \emph{random utility model} \cite{thurstone1927, marschak1959, luce1959}.
% TODO: figure only if there is enough space
% Given a utility function, it is straightforward to define top-1 choices
% (see \cref{fig:utility1d}).
% % TODO: formal definition (?)
% \begin{figure}[tb]
% 	\centering
% 	\includegraphics[width=0.71\linewidth]{pics/viz_utility1d}
% 	\caption{A set of objects is mapped into a 1-dimensional utility space.}
% 	\label{fig:utility1d}
% \end{figure}

At first glance, it may appear that this approach can easily be generalized to modeling subset
choices:
Instead of only selecting the top-1 object, one could consider to select the
top-$k$ objects, where $1 \leq k \leq |Q|$.
One major drawback of this approach is that the subset size is predetermined to
be $k$, so it is not possible to produce subsets of varying size.
Another possibility is to specify a threshold for the utilities, and to include all
objects with a utility higher than the threshold in the
choice set \cite{pfannschmidt2019}.
While this allows for the prediction of subsets of arbitrary size, the decision of
whether to include an object in the choice set is now completely independent of all
the other objects.

As we shall see in the next section, there is a natural way to define
subset choices, if we embed the objects in a
higher-dimensional utility space.

\section{Pareto-Embeddings}
\label{sec:pareto_embedding}

\begin{figure}[tb]
  \centering
  \includegraphics[width=0.6\linewidth]{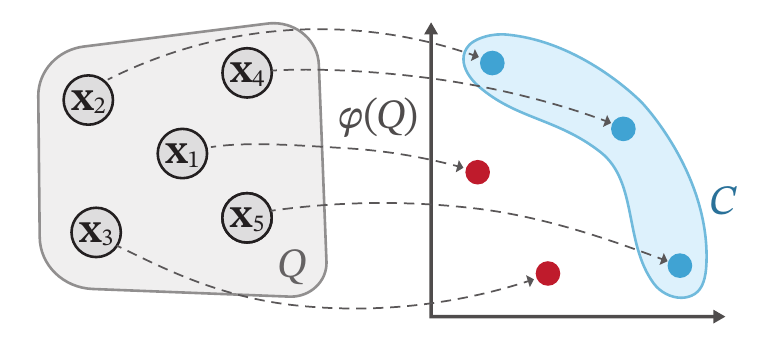}
  \caption{A Pareto-embedding $\varphi(\cdot)$ maps a given set of objects $Q$ into a
    higher-dimensional space $\cZ$.
    The Pareto-optimal points in this space we
    define to be the choice set $C$.
  }
  \label{fig:pareto_embedding}
\end{figure}

The basic idea of a Pareto-embedding is illustrated in \cref{fig:pareto_embedding}.
On the left side, we depict the original set of objects in the object
space $\cX$.
The function $\varphi \from \cX \to \cZ$ maps each point into a new embedding
space $\cZ \subseteq \IR^{d'}$.
This space can be thought of as a higher-dimensional utility space,
\ie each dimension corresponds to the utility on a certain aspect or criterion.
% relevant to the decision maker. 
%which the decision maker wants to maximize.
As we can see, the choice set $C$ forms what is called a Pareto-set in this
new space, \ie the set of points that are not dominated by any other
point.

More formally, let $Q = \{\vec{x}_1, \dots, \vec{x}_m\} \subseteq \mathcal{X}$ be the original set of objects and
$Z = \varphi(Q) =  \{\vec{z}_1, \dots, \vec{z}_m\} =  \{\varphi(\vec{x}_1), \dots, \varphi(\vec{x}_m)\} $ the corresponding points in the
embedding space.
A point $\vec{z}_i$ in the embedding space is \emph{dominated} by another point $\vec{z}_j$ if $z_{i, k} \leq z_{j, k}$ for all $k \in [d']$ and $z_{i, k} < z_{j, k}$ for at least one such $k$.
Then, a point $\vec{z}_i$ is called \emph{Pareto-optimal}
(with respect to $Z$), if it is not dominated by any other point $\vec{z}_j \in Z$, $1 \leq j\neq i \leq m$.
We denote by $P_\varphi(Q) \subseteq Q$ the subset of points that are Pareto-optimal in $Q$ under the mapping $\varphi$, i.e., the points $\vec{x}_i \in Q$ such that $\varphi(\vec{x}_i)$ not dominated by any point in $\{\varphi(\vec{x}_1), \dots, \varphi(\vec{x}_m)\} $.
%there exists a $k \in [1, d']$ such that $z_{i, k} > z_{j, k}$.
%\begin{equation}\label{eq:paretoopt}
%  \forall \vec{z}_j \in \varphi(Q), 1 \leq j \neq i \leq n\
%  \exists k \in [1, d']:\
%  z_{i, k} > z_{j, k}
%\end{equation}
% TODO: For consistency, should we call this Pareto-dominated?

%\begin{equation}\label{eq:dominated}
%  \exists \vec{z}_j \in \varphi(Q), 1 \leq j \neq i \leq n\
%  \forall k \in [1, d']:\
%  z_{i, k} < z_{j, k}
%\end{equation}

%\begin{definition}
%  Given an object space $\cX$ and a choice space $\cC$,
%  we call $(\varphi, \cZ)$ with
%  $
%    \varphi \from \cX \to \cZ
%  $
%  a \emph{Pareto-embedding} if for all $Q \subseteq \cX$ and $C \in \cC$
%  with $C \subseteq Q$, it holds that:
%  \begin{enumerate}
%    \item For all $\vec{x} \in C$: $\varphi(x)$ is \emph{Pareto-optimal} in set $\varphi(Q)$.
%    \item For all $\vec{x} \in Q\setminus C$: $\varphi(x)$ is \emph{dominated} in set $\varphi(Q)$.
%  \end{enumerate}
%\end{definition}
%As is apparent, as soon as we have a Pareto-embedding $\varphi$, it becomes trivial to
%predict the chosen subset.
It is interesting to note that the traditional one-dimensional utility always imposes a total order relation on
the available objects, whereas the Pareto-embedding generalizes this to a partial order.
% TODO: This sentence reads a bit "over the top"
Therefore, richer preference structures with multiple layers of incomparability can be modeled.

% TODO: Mention multi-self somewhere
% TODO: Transition to learning part

%\subsection{Learning a Pareto-Embedding}
%\label{sub:learning}

%As is apparent, the crucial step is to learn an embedding function $\varphi$, which satisfies the previously discussed constraints, since the actual step of predicting a subset is now straightforward.

Given a set of observed choices $\mathcal{D} = \{ (Q_n , C_n) \}_{n=1}^N$ as training data, where $Q_n \subseteq \mathcal{X}$ is a choice task and $C_n \subseteq Q_n$ the subset of objects selected, we are interested in learning a Pareto-embedding $\varphi$ coherent with this data in the sense that $C_n \approx P_\varphi(Q_n)$ for all $n \in [N]$.
Obviously, a function of that kind can then also be used for predictive purposes, i.e., to predict the choice for a new choice task.
To induce $\varphi$ from $\mathcal{D}$, we devise a general-purpose loss function, which can be used
with any end-to-end trainable model, and hence should be differentiable almost everywhere.
% TODO: More on the aspect of "unbiasedness" wrt the embedding 
%We will now introduce the loss terms $L_{\text{PO}}$ and $L_{\text{DOM}}$.

%Introducing both terms formally and explain them using \cref{fig:loss}.

The loss function we propose consists of several components, which we introduce step by step.
Consider a choice $C$ in a choice task $Q$, and denote by $\vec{c} \in \{0,1\}^{|Q|}$ the vector encoding of $C$, i.e., $c_i = 1$ if $\vec{x}_i \in C$ and $c_i = 0$ otherwise.
In order to accomplish $C = P_\varphi(Q)$,
the first constraint to be fulfilled by $\varphi$ is to ensure that
each point $\vec{x}_j \in C$ will have an image in the embedding space
which is Pareto-optimal in $Z$.
\begin{figure}[tb]
  \centering
  \subfloat[$L_{\text{PO}}$\label{fig:loss1}]{
    \includegraphics[width=0.3\textwidth]{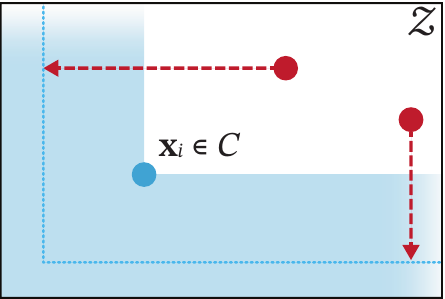}
  }
  \hfil
  \subfloat[$L_{\text{DOM}}$\label{fig:loss2}]{
    \includegraphics[width=0.3\textwidth]{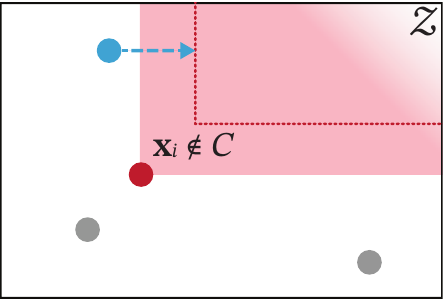}
  }
  \caption{Visualization of the effect of the loss terms $L_{\text{PO}}$ and $L_{\text{DOM}}$ in $\cZ$ space.}
  \label{fig:loss}
\end{figure}
Consider \cref{fig:loss1}, where the point in blue depicts the
image $\vec{z}_j = (z_{j,1}, \ldots , z_{j,d'})$ of $\vec{x}_j$.
The loss needs to penalize all points dominating $\vec{z}_j$ (shown in red).
Formally, the first part of the loss function is defined as follows:
\begin{align}
  L_{\text{PO}}(Z, \vec{c}) & = \sum_{1 \leq i \neq j \leq |Z|}
  \max\Bigl( 0, c_j\cdot \min_{1\leq k \leq d'} (1 + z_{i,k} - z_{j,k}) \Bigr)
  \label{eq:po}
\end{align}
%Here $\indic{\cdot}$ is the indicator function, \ie $\indic{P}=1$ if the predicate $P$ is true and $=0$ otherwise.
We project the points towards the blue region using the minimum term and
penalize them in proportion from their distance to the boundary.
Note that, to enforce a margin effect, we already penalize non-dominating points close to the boundary.
This corresponds to using a hinge loss upper bound on
the 0/1-binary loss, which is 1 if $\vec{z}_i$
dominates $\vec{z}_j$ and 0 otherwise.

Similarly, we define a loss that penalizes the embedding of a point
$\vec{x}_i \in Q\setminus C$ so that $\vec{x}_i$ is not dominated:
\begin{align}
  L_{\text{DOM}}(Z, \vec{c}) & =
  \sum_{i=1}^{\abs{Z}} (1 - c_i)
  \min_{ j\neq i}
  \biggl( \, \sum_{k=1}^{d'} \max \big(0, 1+ z_{i,k}- z_{j,k} \big) \biggr)
  \label{eq:dom}
\end{align}
The minimum selects the point which is closest to dominating
$\vec{z}_i$, while the inner sum penalizes all dimensions in
which this point is not yet better than $\vec{z}_i$.

With these two terms, we can ensure that if the loss is 0, we have a valid Pareto-embedding of the points.
Furthermore, we add two more terms that are useful.
To preserve as much of the original structure present in
the object space $\cX$, we use multidimensional scaling (MDS) \cite{mead1992}.
It ensures that objects close to each other in the object space
$\cX$ will also be close in the embedding space $\cZ$.
In addition, all the losses so far are shift-invariant in the
embedding space.
To make the solution identifiable, we regularize the mapped points
towards 0 using an $L_2$ loss.
We define the complete Pareto-embedding loss as a convex combination
%($\norm{\vec{\alpha}}_1 = 1$) of all independent loss terms:
\begin{align*}
  L(Q, Z, \vec{c}) & =
  \alpha_1 L_{\text{PO}}(Z, \vec{c})
  + \alpha_2 L_{\text{DOM}}(Z, \vec{c})
  + \alpha_3 L_{\text{MDS}}(Q, Z)
  + \alpha_4 \sum_{i=1}^{\abs{Z}}\norm{\vec{z}_i}_2
  %\label{eq:loss}
\end{align*}
with weights $\alpha_1, \alpha_2, \alpha_3, \alpha_4 \geq 0$ such that $\alpha_1 + \alpha_2 + \alpha_3 + \alpha_4 = 1$.
These weights can be treated as hyperparameters of the learning algorithm.
Given a space $\Phi$ of embedding functions, this algorithm seeks to find a minimizer
\[
  \varphi^* \in \operatorname*{argmin}_{\varphi \in \Phi} \sum_{n=1}^N L \big(Q_n, P_\varphi(Q_n), \vec{c}_n \big)
  %and can be optimized during validation time.
\]
of the overall loss on the training data $\mathcal{D}$.

%\subsection{A Learning Algorithm for Pareto-Embeddings}
\newpage
\section{Empirical Evaluation}\label{sec:evaluation}

\begin{figure}[tb]
  \centering
  \includegraphics[width=0.6\linewidth]{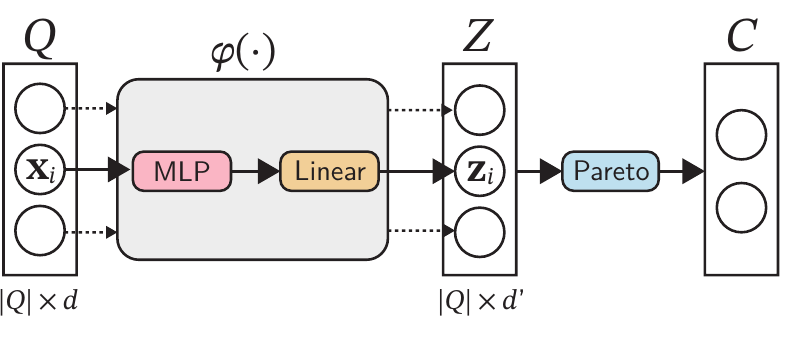}
  \caption{Architecture of our approach.
    Each object is passed through a
    (deep) multi-layer perceptron followed by a linear output layer to produce the embedding.
    The Pareto-optimal points are selected to obtain a prediction $C$.
  }
  \label{fig:architecture}
\end{figure}

%With a suitable loss function at hand,
As for the space of embedding functions, any model class amenable to training by gradient descent can in principle be used.
Here, as a proof of concept, we use a simple fully connected multi-layer perceptron
as a learner.
The architecture is depicted in \cref{fig:architecture}.
We take each object $\vec{x}_i$ for $1\leq i\leq |Q|$ of the task $Q$ and pass it through the
(deep) multi-layer perceptron.
Rectified linear units are used here as the nonlinearities.
Batch normalization \cite{Ioffe2015} is applied after each layer to speed up and stabilize training.
In the final layer, we pass the output of the multi-layer perceptron through a linear layer
with $d'$ outputs.
After the same network (using weight sharing) was applied to all objects in $Q$,
we end up with the transformed set $Z$.
To obtain the final prediction, we take the set $Z$ and compute the corresponding
Pareto-set.
The network can be trained using standard backpropagation of the loss.

To ascertain the feasibility of learning a Pareto-embedding from data,
we evaluate our approach on a suite of benchmark problems from the field of multi-criteria optimization.
We use the well-known DTLZ test suite by \citet{Deb2005} and the ZDT test suite
by \citet{Zitzler2000}, containing datasets of varying difficulty.
Adding a simple two-dimensional two parabola (TP) dataset, we end up with 14 benchmark problems in total.
We generate \num{40960} object sets of size 10 with 6 features each for every problem.
Exceptions are the TP dataset with only 2 features and the ZDT5 dataset, which
has 35 binary features by definition.
For the DTLZ problems, we set the dimensionality of the underlying objective space to 5.
%we can choose the dimensionality of the underlying objective space,
%which we set to 5.

We evaluate our approach on every problem by 5 repetitions of a Monte Carlo cross validation with a 90/10\% split into training and test data.
%-fold (shuffle) cross validation, where we split off \SI{10}{\percent} test instances each fold.
The remaining instances are split into \si{1/9} validation instances and
\si{8/9} training instances.
We use the validation set to jointly optimize the hyperparameters of the
learner, which are
\begin{inparaenum}[(a)]
  \item the loss weights $\alpha_1,\alpha_2, \alpha_3, \alpha_4$,
  \item the maximum learning rate of the cyclical learning rate scheduler, and
  \item the number of hidden units and layers,
\end{inparaenum}
using \num{60} iterations of Bayesian optimization.
The neural network was trained for 500 epochs.
The number of embedding dimensions $d'$ we set to 2, since this allows
us to move from a total order (only one utility dimension) to a partial order.

Finally, we need a suitable metric to compare the ground truth subsets to the
predicted ones.
Since the shape of the Pareto-sets has an impact on how many points end up in
the chosen subset, we have varying levels of positives across the datasets.
Therefore, we choose a metric that is unbiased with respect to the prevalence
of positives and well-suited for problems with class imbalance, called the
\emph{A-mean} \cite{menon2013}, the arithmetic mean of the true positive and
true negative rate.

\begin{figure}[tb]
  \centering
  \includegraphics[width=\linewidth]{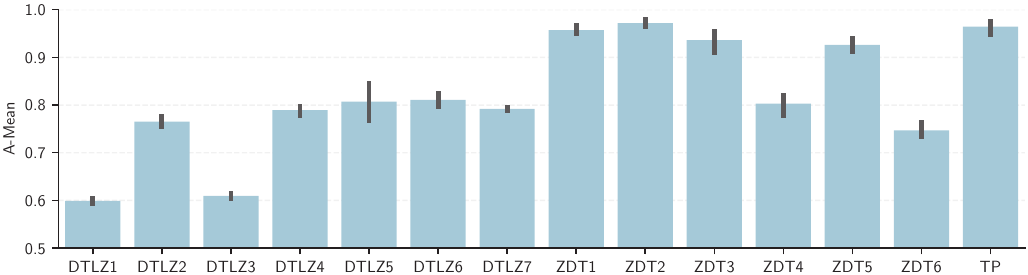}
  \caption{Results of the empirical evaluation.
    The bars show the average
    performance in terms of A-mean across the 5 outer splits.
    The sticks show the estimated standard deviation.
  }
  \label{fig:results}
\end{figure}

The results %in terms of the A-mean 
are shown in \cref{fig:results}.
For five of the problems, the embedding approach is able to achieve an
average A-mean of over \SI{90}{\percent}, indicating that for these
problems we often identify the choice set correctly.
This is important, as it shows that the loss function is able
to steer the model parameters towards a valid Pareto-embedding.
For comparison, a random selection in which each object is included in the choice set with a fixed probability (independently of the others) achieves an average A-mean of \SI{50}{\percent}.
Thus, it is apparent that our learner is performing better than random guessing
on all datasets.
%It is apparent, that the DTLZ problems are on average harder to
%learn than the ZDT problems
We also did an ablation experiment, where we removed the MDS term from the
loss function and repeated the complete training procedure (including optimization of
all the other hyperparameters).
This resulted in a significant decrease in performance, showing that the MDS term
is not only useful to preserve distances, but adds a helpful inductive bias.

\section{Conclusion and Outlook}

We proposed a novel way to tackle the problem of learning choice functions.
Viewing it as an embedding problem and transforming the given objects into a utility space of more than one dimension,
subset choice are naturally identified by the criterion of Pareto-optimality.
To learn an embedding from a given set of observed choices as training data, we developed a suitable loss function that penalizes violations of the Pareto condition.
Encouraged by the promising first results on benchmark problems, we are now
looking forward to a more extensive empirical evaluation and applications to real-world choice problems.
%compare the approach to existing algorithms.

%
%\subsubsection*{Acknowledgments}
% TODO: We need to acknowledge the PC², since we did computation on the OCuLUS:
% The authors gratefully acknowledge the funding of this project by computing time provided by the Paderborn Center for Parallel Computing (PC²).
%
%
% ---- Bibliography ----
%
%\newpage
\renewcommand{\bibsection}{\section*{References}} % required for natbib to have "References" printed and as section*, not chapter*
\bibliographystyle{splncsnat}

\begingroup
\microtypecontext{expansion=sloppy}
\small % ensure correct font size for the bibliography
\bibliography{references.bib}
\endgroup
\end{document}